\documentclass[runningheads]{llncs}
\usepackage{graphicx}
\usepackage{xcolor}
\usepackage{array}
\RequirePackage{algorithm, algorithmicx, algpseudocode}
\begin{document}
\title{Semi-Supervised Deep Learning Using Improved Unsupervised Discriminant Projection\thanks{This work is supported by NSFC China (61806125, 61802247, 61876107) and Startup Fund for Youngman Research at SJTU (SFYR at SJTU). Authors with * make equal contributions. Corresponding author: Enmei Tu (hellotem@hotmail.com)}}
\titlerunning{Semi-supervised Deep Learning Using UDP}
\author{Xiao Han$^*$ \and
	Zihao Wang$^*$ \and
Enmei Tu \and
Gunnam Suryanarayana \and
Jie Yang}
\authorrunning{Xiao Han et al.}
\institute{School of Electronics, Information and Electrical Engineering,\\
	Shanghai Jiao Tong University, China\\
\email{hanxiao2015@sjtu.edu.cn, wangzihao33@sjtu.edu.cn, hellotem@hotmail.com}}
\maketitle              % typeset the header of the contribution
\begin{abstract}

Deep learning demands a huge amount of well-labeled data to train the network parameters.  How to use the least amount of labeled data to obtain the desired classification accuracy is of great practical significance, because for many real-world applications (such as medical diagnosis),  it is difficult to obtain so many labeled samples. In this paper, modify the unsupervised discriminant projection algorithm from dimension reduction and apply it as a regularization term to propose a new semi-supervised deep learning algorithm, which is able to utilize both the local and nonlocal distribution of abundant unlabeled samples to improve classification performance. Experiments show that given dozens of labeled samples,  the proposed algorithm can train a deep network to attain satisfactory classification results.

\end{abstract}
\section{Introduction}

In reality, one of the main difficulties faced by many machine learning tasks is manually tagging large amounts of data. 
This is especially prominent for deep learning, which usually demands a huge number of well-labeled samples.
 Therefore, how to use
the least amount of labeled data to train a deep network has become an important topic in the area. To overcome this problem, researchers proposed that the use of
a large number of unlabeled data can extract the topology of the overall
data's distribution. Combined with a small amount of labeled data, the
generalization ability of the model can be significantly improved, which is
the so-called semi-supervised learning \cite{chapelle2009semi,zhu2009introduction,weston2012deep}.

Recently, semi-supervised deep learning has made some progress. The main ideas of existing works broadly fall into  two categories. One is generative model based algorithms, for which unlabeled samples help the generative model to learn the underly sample distribution for sample generation. Examples of this type algorithms include CatGAN \cite{springenberg2015unsupervised}, BadGAN \cite{dai2017good}, variational Bayesian \cite{kingma2014semi}, etc. 
The other is discriminant model based algorithms, for which the role of the unlabeled data  may provide sample distribution information to prevent model overfitting , or to make the model more
resistant to disturbances. Typical algorithms of this type include unsupervised loss regularization \cite{thulasidasan2016semi,Bachman2014Learning}, latent feature embedding \cite{weston2012deep,yang2016revisiting,hoffer2016semi,rasmus2015semisupervised}, pseudo label \cite{lee2013pseudo,wu2018semisupervised}. Our method belongs to the
second category, in which an unsupervised regularization term, which captures the local and global sample distribution characteristics,  is added to
the loss function for semi-supervised deep learning.

\begin{figure}[htbp!]
  \centering \includegraphics[width=0.8\textwidth]{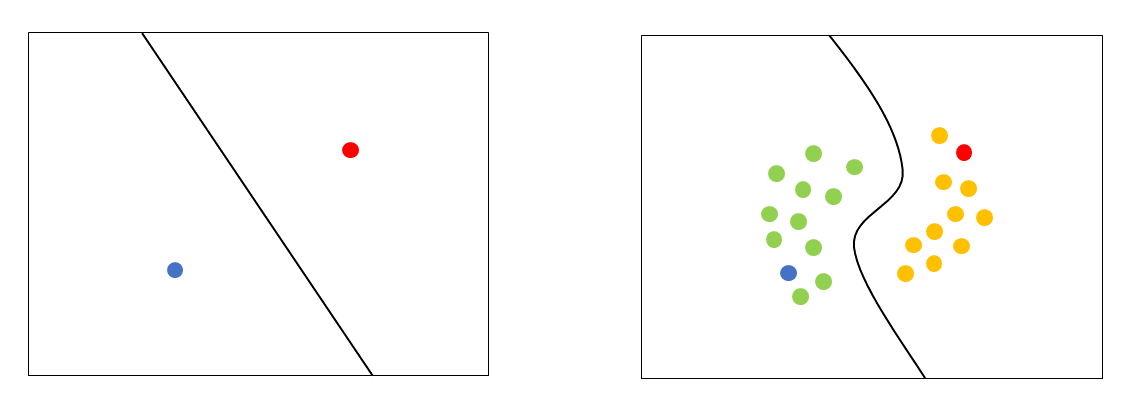}
  \caption{Manifold regularization makes the decision boundary where the
    data distribution is sparse. Left: traditional supervised learning results; right: manifold regularized semi-supervised learning.}
  \label{fig:semi}
\end{figure}

The proposed algorithm is based on the theory of manifold regularization, which is developed by Belkin et
al.\cite{Belkin2006Manifold,belkin2005manifold} and then introduced into deep learning by Weston et al. \cite{weston2012deep}. Given $L$ labeled samples $x_1, x_2,...x_L$ and their corresponding labels $y_1, y_2, ..., y_L$, recall that  manifold regularization combines the idea of
manifold learning with the idea of semi-supervised learning, and learns the
manifold structure of data with a large amount of unlabeled data, which
gets the model better generalization. Compared to the loss function in tradition
supervised learning framework, the manifold regularization based semi-supervised learning algorithm adds a
new regularization term to penalize the complexity of the discriminant function $f$ over the sample distribution manifold, as shown in the equation (\ref{eq:manifold}):

\begin{equation}
  \label{eq:manifold}
  \frac{1}{L}\sum_{i=1}^LV(x_i,y_i,f)+\gamma_A\left\lVert f\right\rVert_K^2+\gamma_I\left\lVert f \right\rVert_I^2
\end{equation}
where $V(\cdot)$ is an arbitrary supervised loss term, and $\left\| {\cdot} \right\|_K$ is a
kernel norm, such as a Gaussian kernel function, that penalizes the model complexity in the ambient (data) space. $\left\| {\cdot} \right\|_I$ is the introduced
manifold regularization term, which penalizes model complexity along the data distribution manifold to make sure that the  prediction output have
the same distribution as the input data. $\gamma_A$ and $\gamma_I$ are used
as weights. As shown in Fig. \ref{fig:semi}, after the manifold
regularization term is introduced, the decision boundary tries not to
destroy the manifold structure of the data distribution and meanwhile, keeps itself as simple as possible, so that the
boundary finally passes through where the data is sparsely distributed.

However, the research on the application of manifold regularization in
the field of semi-supervised deep learning has not been fully explored. The
construction of manifold regularization only considers the local structural
relationship of samples. For classification problems, we should
not only preserve the positional relationship of neighbor data to ensure
clustering, but also consider distinguishing data from different manifolds
and separating them in the embedded space. Therefore, in this paper, we propose a
novel manifold loss term based on the improved Unsupervised Discriminant
Projection (UDP) \cite{Jian2007Globally}, which incorporates both local and nonlocal distribution information, and we conduct experiments on real-world datasets to demonstrate that it can produce better classification accuracy for semi-supervised deep learning than its counterparts.

The following contents are organized as follows: The theory and the proposed
algorithm are presented in Section 2; then the experimental results are
given in Section 3, followed by conclusions and discussions in Section 4.

\section{Improved UDP Regularization Term}
In this section, we first review the UDP algorithm and then introduce an improved UDP algorithm. Then we propose a semi-supervised deep learning algorithm which is based on the improved UDP algorithm.

\subsection{Basic idea of UDP}

\begin{figure}[htbp!]
  \centering \includegraphics[width=0.7\textwidth]{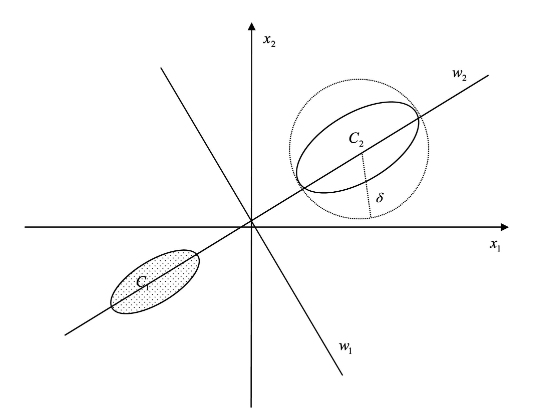}
  \caption{Illustration of clusters of two-dimensional data and optimal
    projection directions \cite{Jian2007Globally}.}
  \label{fig:pro}
\end{figure}

The UDP method is proposed by Yang et al. originally for dimensionality
reduction of small-scale high-dimensional data \cite{Jian2007Globally}. As
a method for multi-manifold learning, UDP considers both local and
non-local quantities of the data distribution. The basic idea of UDP is shown in Fig. \ref{fig:pro}. Suppose that the data is distributed on two elliptical manifolds denoted by $c_1$ and $c_2$, respectively. If we
only require that the distances of neighboring data are still close
after being projected along a certain direction, then the projection along
$\mathbf{w}_1$ will be the optimal direction, but at this time the two data
clusters will be mixed with each other and difficult to separate after projection. Therefore, while
requiring neighbor data to be sufficiently close after projection, we
should also optimize the direction of the projection so that the distance
between different clusters is as far as possible. Such projected data
are more conducive to clustering after dimensionality reduction.

For this reason, UDP uses the ratio of local scatter to non-local scatter,
to find a projection which will draw the close data closer, while
simultaneously making the distant data even more distant from each other.
The local scatter can be characterized by the mean square of the Euclidean
distance between any pair of the projected sample points that are
neighbors. The criteria for judging neighbors can be $K$-nearest neighbors
or $\epsilon$ neighbors. Since the value of $\epsilon$ is difficult to
determine and it may generate an  unconnected graph, the $K$-nearest neighbor criterion is used here to define the weighted adjacency
matrix $\mathbf{H}$ with kernel weighting:

\begin{equation}
  H_{ij}=
    \begin{cases}
      exp(-\left|\lVert x_i-x_j \right|\rVert^2/t) & \text{if $x_j$ is among $K$ nearest neighbors of $x_i$,}\\
        & \text{or $x_i$ is among $K$ nearest neighbors of $x_j$} \\
      0 & \text{otherwise} 
    \end{cases}
\end{equation}

Then given a training set containing $M$ samples $x_1, x_2, ..., x_M$, denote the
local set $U^K=\{(i, j)| x_j \text{ is the neighbor of } x_i\}$. After projecting $x_i$ and $x_j$ onto a direction $\mathbf{w}$, we get their
images $y_i$ and $y_j$. The local scatter is defined as
\begin{equation}
  J_L(\mathbf{w})=\frac{1}{MM}\sum_{i=1}^M\sum_{j=1}^MK_{ij}(y_i-y_j)^2
\end{equation}

Similarly, the nonlocal scatter can be defined by the mean square of the
Euclidean distance between any pair of the projected sample points that
are not in any set of neighborhoods. It is defined
as
\begin{equation}
  J_N(\mathbf{w})=\frac{1}{MM}\sum_{i=1}^M\sum_{j=1}^M(K_{ij}-H_{ij})(y_i-y_j)^2
\end{equation}
The optimal projection vector $\bf w^*$ minimizes the following final objective function
\begin{equation}
{\bf{w}}^*=\arg \min J({\bf{w}})=\frac{J_L}{J_N}
\end{equation}
\subsection{An improved UDP for large scale dimension reduction}
Since the original UDP method is developed for
dimensionality reduction of small-scale data sets, the data outside the
$K$-nearest neighbors of a sample are regarded as nonlocal data and participate in the
calculation of a nonlocal scatter. However, when the scale of training data
is large, this way of calculating the nonlocal scatter will bring a prohibitive
computational burden, because each sample has $M-K$ nonlocal data. To overcome this problem, we propose an improved UDP for large scale dimension reduction.

Suppose there are training data $x_1, x_2, ..., x_M$ ,and the desired output of
$x_i$ after dimension reduction is $y_i$. Using the Euclidean distance
as a measure, similar to the definition of the $K$-nearest neighbor set, we
define a set of $N$-distant data set $D^N=\{(i,j)|x_j \text{ is one of the $N$
  farthest data from } x_i\}$. Similarly, we define a non-adjacency matrix
$\mathbf{W}$:
  \begin{equation}
    W_{ij}=
      \begin{cases}
        exp(-\left|\lVert x_i-x_j \right|\rVert^2/t) & \text{if $x_j$ is among $N$ farthest samples away from $x_i$,}\\
              & \text{or $x_i$ is among $N$ farthest samples away from $x_j$} \\
        0 & \text{otherwise}.
      \end{cases}
\end{equation}
Then we define the distant scatter as

\begin{equation}
  J_D=\frac{1}{m}\sum_{i=1,j\in D^N}^MW_{ij}\left\lVert y_i-y_j \right\rVert_2^2
\end{equation}
for the local scatter $J_L$, we use the same one as the original UDP. So the objective function of the improved UDP is

\begin{align}
  J_R(\bf w)= & \frac{J_L}{J_D}\\
  = & \sum_{i=1}^M\frac{\sum_{j\in U^K}H_{ij}\left\lVert y_i-y_j \right\rVert_2^2}{\sum_{b \in D^N}W_{ib}\left\lVert y_i-y_b \right\rVert_2^2}
\end{align}

The improved UDP also requires that after the mapping of
the deep network, the outputs of similar data is as close as possible,
while simultaneously ``pushing away'' the output of dissimilar data. Although
only the data with extreme distance is used, in the process of making the
dissimilar data far away from each other, the data similar to them will
gather around them respectively, thus widening the distance between the
classes and making the sparse area of data distribution more sparse,
densely areas denser.
\subsection{The improved UDP based semi-supervised deep learning}
Suppose we have a dataset $\{x_1, ..., x_L, x_{L+1},...x_M\}$, in which the first $L$ data points are labeled samples with labels  $\{y_1, y_2, ..., y_L\}$, and the rest data points are unlabeled samples. Let $\{g_1, g_2, ..., g_M\}$ be the embeddings of the samples through a deep network.  Our aim is to train a deep network $f(x)$ using both labeled and unlabeled samples, such that different classes are well separated and meanwhile, cluster structures are well preserved. Putting all together, we have the following objective function
\begin{equation}
  J=\sum_{i=1}^Ll(f_i,y_i)+\lambda\sum_{i=1,j \in U^K, k \in D^N }^{L+U}J_R(g_i, g_j, g_k, H_{ij}, W_{ik})
  \label{SSL_Obj}
\end{equation}
where $L$ is the number of labeled data and $U$ is the number of unlabeled
data. $l(\cdot)$ is the supervised loss function and $J_R(\cdot)$ is the UDP
regularization term. $\lambda$ is the hyperparameter, which is used to
balance the supervisory loss and unsupervised loss. We use the softmax function as our supervised loss, but other type of loss function (e.g. mean square error) are also applicable.

We use error backpropagation (BP) to train the network. The details of the training process are given in the following algorithm. 

\begin{algorithm}
  \caption{Semi-supervised deep learning based on improved UDP}
  \label{algo:UDP}
  \begin{algorithmic}[1]
    \Require labeled data $x_i$ and corresponding label $y_i,
    i=1,2,...,L$, unlabeled data $x_j, j = 1,2,...,U$, output of neural
    network $f(\cdot)$, output of the embedded UDP regularization item
    $g(\cdot)$
    
   % \State Calculate the full connection graph of all the data
    
    \State Find $K$-nearest neighbors and $N$-distant samples of each sample

    \State Calculated the kernel weights $H_{ij}$ for neighbors and $W_{ij}$
    for distant samples

    \Repeat

    \State Randomly select a group of labeled data and their labels
    $(x_i,y_i)$

    \State Gradient descend $l(f(x_i),y_i)$

    \State Select $x_i$ and its $K$-nearest data $x_j$ and $N$-distant data
    $x_k$

    \State Gradient descend $J_R(g(x_i),g(x_j),g(x_k), H_{ij}, W_{ik})$

    \Until

    \State Meet accuracy requirements or complete all iterations
  \end{algorithmic}
\end{algorithm}

\section{Experimental Results}
\subsection{Results of dimensionality reduction}
Firstly, we test the dimensionality reduction performance of the improved UDP method in two different image datasets, MNIST and ETH-80\footnote{ETH-80:https://github.com/Kai-Xuan/ETH-80}. Then we compare the improved UDP with original UDP, as well as several popular dimension reduction algorithms (Isomap \cite{balasubramanian2002isomap}, Multidimensional scaling (MDS) \cite{cox2000multidimensional}, t-SNE \cite{maaten2008visualizing} and spectral embedding \cite{luo2003spectral}), to show its performance improvement. 

\begin{figure}[htbp!]
	\centering \includegraphics[width=1.0\textwidth]{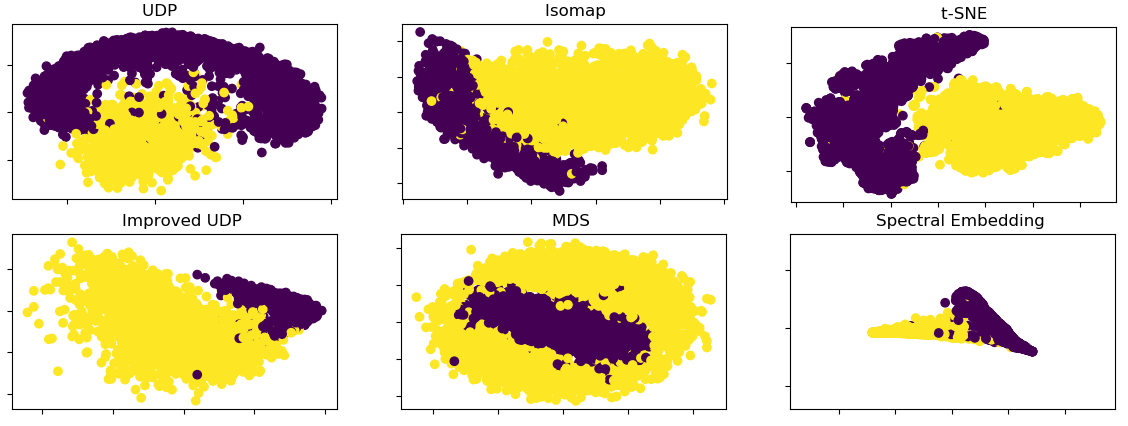}
	\caption{Dimension reduction of digits 1 and 2 in MNIST ($t=4, K=10$ and $N=300$).}
	\label{fig:Mnist}
\end{figure}

\begin{figure}[htbp!]
	\centering \includegraphics[width=1.0\textwidth]{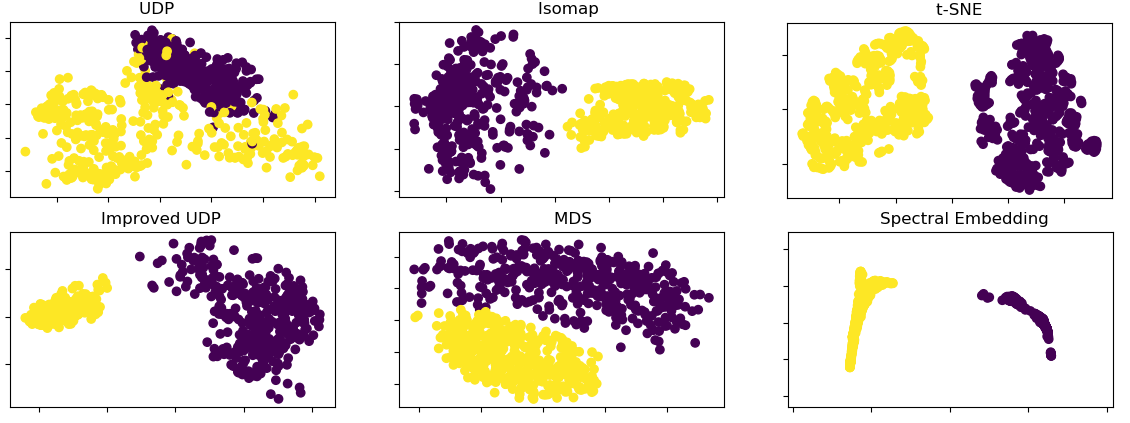}
	\caption{Dimension reduction of `apples' and `pears' categories in ETH-80 ($t=12.0, K=10 $ and $ N=50 $).}
	\label{fig:ETH}
\end{figure}

MNIST is a dataset consisting of $ 28 \times 28$ grayscale images of handwritten digits. We randomly selected 5000 samples from the dataset to perform our experiments because the original UDP usually applies to small-scale datasets. ETH-80 is a small-scale but more challenging dataset which consists of $ 256 \times 256$ RGB images from 8 categories. We use all the 820 samples from ``apples'' and ``pears'' categories and convert the images from RGB into grayscale for manipulation convenience. The parameters of the baseline algorithms are set to their suggested default values  and the parameters (kernel width $t$, number of nearest neighbors $K$ and number of farthest points $N$) of the improved UDP are set empirically. The experimental results on the two datasets are shown in Fig. \ref{fig:Mnist} and Fig. \ref{fig:ETH}, respectively.

From these results we can see that after dimension reduction, the improved UDP maps different classes more separately than the original UDP on both datasets. This is important because while adopting the new UDP into semi-supervised learning in equation  (\ref{SSL_Obj}),  better separation means more accurate classification. It is also worth mentioning that although on the ETH-80 dataset, the improved UDP achieves comparable results as the rest baseline algorithms, its results on MNIST is much better (especially than MDS, Isomap) in terms of classes separation.

To quantitatively measure the classes separation, Table \ref{tab1} shows the cluster purity  given by k-means clustering algorithm on these two datasets after dimensionality reduction. The purity is calculated based maximum matching degree \cite{tu2014novel} after clustering. 
\begin{table}
	\centering
	\caption{Cluster purity of 6 methods.}\label{tab1}
	\begin{tabular}{|p{3.7cm}<{\centering}|p{1.7cm}<{\centering}|p{1.7cm}<{\centering}|}
		\hline 
		Method & MNIST & ETH-80 \\ 
		\hline 
		UDP & 81.7 & 77.7 \\ 
		\hline 
		Improved UDP & \textbf{93.8} & \textbf{99.4} \\ 
		\hline 
		%LLE\cite{roweis2000nonlinear} & 98.4 & 99.9 \\ 
		%\hline 
		%Modified LLE\cite{zhang2007mlle} & 99.1 & 100.0 \\ 
		%\hline 
		Isomap \cite{balasubramanian2002isomap} & 86.1 & 99.9 \\ 
		\hline 
		MDS \cite{cox2000multidimensional} & 53.7 & 98.9 \\ 
		\hline 
		t-SNE \cite{maaten2008visualizing} & 93.1 & 100.0 \\ 
		\hline 
		Spectral Embedding \cite{luo2003spectral} & 98.6 & 100.0 \\ 
		\hline 
	\end{tabular} 
\end{table}

Table \ref{tab1} demonstrates that our improved UDP method improves the cluster purity by a large margin compared to the original UDP. It can also be seen from Fig. \ref{fig:Mnist} and Fig. \ref{fig:ETH} that our improved UDP method is more appropriate for clustering than original UDP. Furthermore, our method is more efficient than the original UDP because we do not have to calculate a fully connected $ M \times M $ graph. What we need are the kernel weights of the $ K $ neighbors and $ N $ distant data. On both datasets, our method gets much better (on MNIST) or competitive results with other dimension reduction methods.

\subsection{Results of classification}
We conduct experiments on MNIST dataset and SVHN dataset\footnote{SVHN the 
The Street View House Numbers (SVHN) Dataset (http://ufldl.stanford.edu/housenumbers/), which consists of color images for real-world house number digits with various appearance and is a highly challenging classification problem.} to compare the proposed algorithm with the supervised deep learning (SDL) and Manifold Regularization (MR) semi-supervised deep learning \cite{weston2012deep}. The number of labeled data for MNIST dataset is set to 100, combined with 2000 unlabeled data, to train a deep network. For SVHN, from the training set we randomly selected 1000 samples as labeled data and 20000 samples as unlabeled data  to train a network. For both experiments, we test the trained network on the testing set (of size 10000 in MNIST and 26032 in SVHN ) to obtain testing accuracy.
The optimizer we choose Adam. The parameters are manually tuned using a simple grid search rule. $K$ and $N$ take 10 and 50 and kernel width is within $[3.5,4.0]$.

 We adopt the three embedding network structures described in \cite{weston2012deep} and the results of MNIST and SVHN are shown in Table \ref{tab2}. For supervised deep learning, we  apply entropy loss at the network output layer only, since middle layer embedding and auxiliary network do not make any sense. From the table we can see, MR is better for middle layer embedding. Our method is better for output embedding and auxiliary network embedding and  achieves better classification results for most network structures. The results also suggest that it may be helpful to combine MR with UDP together, using MR for hidden layer and UDP for output layer\footnote{We leave this to our future work. We should also point out  that although the classification accuracies are somehow lower than the state-of-the-art results, the network we employed is a traditional multilayer feedforward network and we do not utilize any advanced training techniques such as batch-normalization, random data augmentation. In the future, we will try to train a more complex network with advanced training techniques to make thorough comparisons.}. 

\begin{table}
  \centering
  \caption{Classification correct rate.}\label{tab2}
%  \resizebox{\textwidth}{!}{
  \begin{tabular}{|c|c|c|c|c|c|c|}
    \hline
   number of labled data &  \multicolumn{3}{c|}{MNIST} & \multicolumn{3}{c|}{SVHN} \\
    \hline
    \multicolumn{1}{|l|}{} & SDL & MR    & Improved UDP   & SDL   &  MR    & Improved UDP     \\
    \hline
    Output layer embedding & 74.31 &  82.95 &\textbf{83.19  }    &  55.21  & 64.70 & \textbf{72.66}    \\
    Middle layer embedding & -& \textbf{83.52} & 83.07          & - &\textbf{72.10} & 69.35         \\
    Auxiliary neural network & - & 87.55 & \textbf{87.79} & - &  62.61 & \textbf{71.32}      \\

    \hline
  \end{tabular}
%  }
\end{table}

\section{Conclusions and Future Work}

Training a deep network using a small number of labeled samples is of great practical significance, since many real-world applications have big difficulties to collect enough labeled samples. In this paper, we modify the  unsupervised discriminant projection (UDP)
algorithm to make it suitable for large data dimension reduction and semi-supervised learning. The new algorithm simultaneously takes both local and nonlocal manifold information into account and meanwhile, could reduce the computational cost.  Based on this, we proposed a new semi-supervised deep learning algorithm to train a deep network with a very small amount of labeled samples and many unlabeled samples.   The
experimental results on different real-world datasets demonstrate its validity and effectiveness. 

The construction of the neighbor graph is based on Euclidean
distance in data space, which may not be a proper distance measure on data manifold. In the future, other neighbor graph construction methods, such
as the measure on the Riemannian manifold, will be tried. The limitation of the current method is that it can attain good results for tasks that are not too complex, such as MNIST, but for more challenging classification datasets, such as CIFAR10, which the direct $K$ nearest neighbors may not reflect the actual similarity, the
method may not perform very well. Our future work will try to use some pre-learning techniques, such as auto-encoder or kernel method, to map origin data to a much concise representation.

\bibliographystyle{splncs04}
\bibliography{cite}
\end{document}